\documentclass[journal]{IEEEtran}
\usepackage{lipsum}
\usepackage{multicol,multirow}
\usepackage[numbers,sort&compress]{natbib} 

\usepackage{tabularx,booktabs}
\usepackage{caption}
\usepackage{subcaption}
\usepackage{tikz-network}
\usepackage{tikz}
\usepackage{tikz-cd}
\usepackage{mathtools}

\usepackage{xspace}
\usepackage{algorithm}
\usepackage{algorithmic}

\usepackage[inkscapearea=page]{svg}
\setsvg{inkscape=inkscape -z -D,svgpath=fig/}

\graphicspath{{./Images/}{./Figures/result/}{./Images/result/}{./images/}}

\usetikzlibrary{angles}

\definecolor{blueish}{RGB}{224,218,242}
\definecolor{yellowish}{RGB}{254,255,165}

\ifCLASSINFOpdf
\else
\fi

\usepackage{amsmath,amssymb} 
\usepackage{url}
\begin{document}
%
\title{Voxel-based 3D Facies Segmentation from Seismic Data: A Comparative Study}
%
%
%

\author{Duc-Thanh Pham, Minh-Tan~Pham, Anh Nguyen, and Van Nguyen
\thanks{D.-T. Pham and V. Nguyen are with the FPT Software AI Center, Hanoi, Vietnam. 
M.-T. Pham is with the Institut de Recherche en Informatique et Systèmes Aléatoires (IRISA), UMR 6074, Université Bretagne Sud, 56000 Vannes, France. A. Nguyen is with the Department of Computer Science, University of Liverpool, Liverpool, United Kingdom. \\
}
}



\maketitle

\begin{abstract}


Seismic facies segmentation has emerged as a significant challenge in geophysics, requiring robust methods and systems to effectively identify geologically analogous facies with limited labeled data. Although existing studies have shown promising results in 2D facies segmentation, they often preprocess the original 3D seismic volumes into sets of 2D slices, typically the inline and crossline directions, and treat this problem as a purely 2D segmentation task. This simplification introduces discontinuities across slices and fails to preserve the spatial and structural continuity in 3D seismic data, thus limiting the model's ability to learn coherent geological patterns. In this work, we present a comparative and reproducible benchmark for voxel-based 3D seismic facies segmentation, built upon publicly available seismic volumes including the Netherlands F3 and the Parihaka datasets, with standardized data splits and evaluation metrics. By evaluating the three representative families of modern 3D segmentation architectures, we establish strong baseline results that highlight the potential and remaining challenges for future research in this domain. 

\end{abstract}

\begin{IEEEkeywords}
Deep learning, 3D imaging, voxel segmentation, seismic imagery, facies segmentation
\end{IEEEkeywords}

%
\IEEEpeerreviewmaketitle

\section{Introduction}
\label{sec:intro}

Seismic facies segmentation has emerged as a significant research focus due to its critical role in addressing key challenges in seismic interpretation for resource exploration, reservoir characterization and subsurface infrastructure development.
Seismic facies are commonly characterized based on distinct seismic attributes such as amplitude, frequency content, waveform shape, and reflector geometry and continuity \citep{Sheriff-1976}. Conventional facies identification frameworks rely heavily on manual analysis of seismic images, requiring experienced interpreters and domain experts to delineate detailed facies \citep{1st_class_facies_classification}. Such a process is not only labor-intensive but also time-consuming. In addition, manual interpretation is often subjective, depending on the interpreter’s experience and skill, limiting its practicality for real-world applications. Over the past decade, the rapid evolution of deep learning (DL) has enabled the automatic learning and recognition of patterns from seismic data, significantly reducing the cost of human interpretation. Numerous DL-based seismic facies classification and segmentation methods have been proposed and shown promising results in both accuracy and inference time (\cite{Civitarese2018_seismicDL}, \cite{Brelaz2022_preSaltCoreDL}, \cite{Zhang2023_ultraDeepCarbonateCNN},  \cite{Fomel2023_deepfacies}). For a comprehensive overview, we refer readers to a recent survey and benchmarking study related to 2D seismic segmentation \cite{principled_benchmark_seismic_segmentation}.

Despite the inherently 3D volumetric nature of seismic data, existing seismic facies segmentation studies have relied on 2D deep learning models, particularly CNN-based architectures such as SegNet, U-Net and DeepLabv3 \cite{principled_benchmark_seismic_segmentation}. This choice is mainly motivated by practical constraints and limited data availability, rather than by the suitability of 2D models for the task itself. Using this strategy, 
seismic cubes are often decomposed into individual 2D slices (such as inlines, crosslines, or time/depth sections), which are processed independently. As a result, these approaches ignore the intrinsic volumetric continuity of seismic data, leading to the loss of inter-slice spatial correlations and potentially producing discontinuous facies patterns across adjacent inlines and crosslines. 
To partially mitigate the limitations of purely slice-based strategies, some studies have explored hybrid solutions that attempt to incorporate additional spatial context while retaining 2D learning frameworks. For example, the authors of \cite{guazzelli2020efficient} proposed an approach that applies 2D CNNs to multiple orthogonal planes (inline, crossline, and time slices), followed by a fusion step to approximate 3D semantic segmentation. While this method improves consistency compared to purely 2D segmentation methods \cite{guazzelli2020efficient}, it still relies on independent 2D feature extraction and requires heuristic fusion to reconstruct the final 3D facies volumes.
Very recently, Tian et al. \cite{tian2025enhancing} have proposed a 2D hybrid framework, where recurrent neural networks (RNNs) are exploited to model sequential dependencies between neighboring slices. By incorporating spatio-temporal relationships, this approach aims to improve inter-slice continuity and reduce discontinuities in the reconstructed facies cubes. However, despite these improvements, \cite{tian2025enhancing} remains fundamentally slice-based, as spatial dependencies are modeled sequentially rather than learned directly from volumetric voxel neighborhoods.
To the best of our knowledge, there exists no prior study that explicitely exploits fully voxel-based DL architectures for seismic facies segmentation to establish standardized and reproducible baselines across multiple public seismic datasets. As a result, the potential benefits of volumetric segmentation models remain remain insufficiently explored. 


To address the gap, this paper presents the first comparative benchmark for 3D seismic facies segmentation using the three representative voxel-based DL architectures including the CNN-based 3D-UNet \cite{3D_UNET}, the transformer-based UNETR \cite{unetr}, and the State-space-model-based SegMamba \cite{segmamba}. We conduct extensive experiments on two widely-used public seismic datasets including the Netherlands F3 \cite{alaudah2019benchmark} and the Parihaka \cite{parihaka}. Following the recently well-established benchmark protocol for 2D seismic facies segmentation \cite{principled_benchmark_seismic_segmentation}, we perform standardized data splits to promote reproducibility and comparability for future studies. 
By releasing our code, data splits and trained models, we aim to establish a reproducible and extensible baseline for 3D seismic facies segmentation research. Our framework not only facilitates fair benchmarking of emerging 3D segmentation methods but also encourages the geoscience and machine learning communities to collaboratively advance toward volumetric challenges of seismic data.

\section{Methodology}

\label{sec:method}
This section describes the datasets, model architectures, evaluation metrics, and implementation details used in our benchmark.
We focus on reproducibility and comparability so that future studies can directly build upon our experimental settings.

\subsection{Datasets and Metrics}
\label{sec:dataset_metric}
We conduct experiments on two publicly available 3D seismic datasets that are widely used in facies interpretation research: the Netherlands F3 Block \cite{alaudah2019benchmark} and the Parihaka 3D survey \cite{parihaka}. Both datasets contain full seismic cubes with facies labels derived from interpreted horizons or well-based stratigraphic units.
\paragraph{F3 Netherlands dataset}

\begin{figure}[ht]
  \centering
  \begin{subfigure}{0.5\textwidth}
    \centering
    \includegraphics[width=\linewidth]{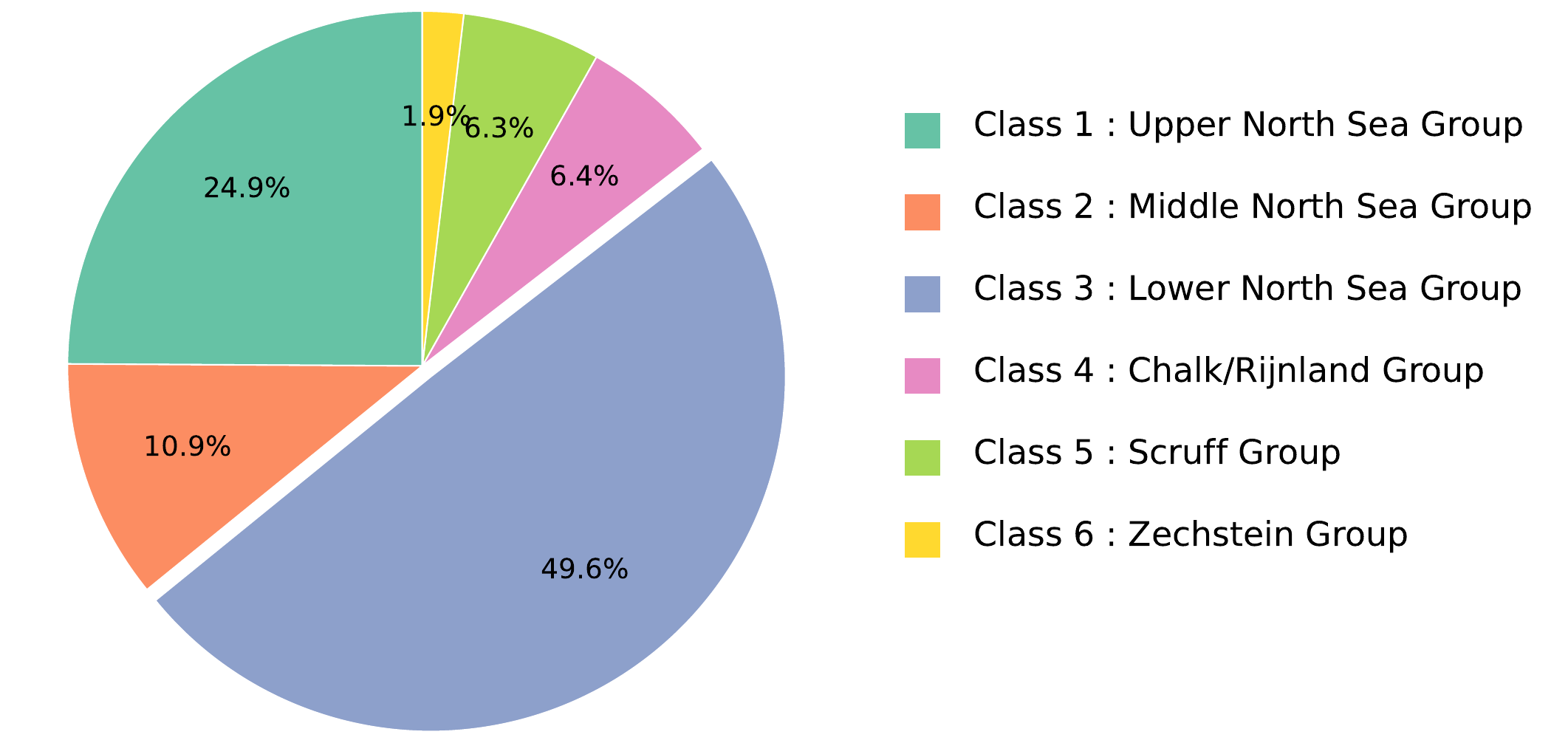}
    \caption{F3 Netherland dataset}
  \end{subfigure}
  \vfill
  \begin{subfigure}{0.5\textwidth}
    \centering
    \includegraphics[width=\linewidth]{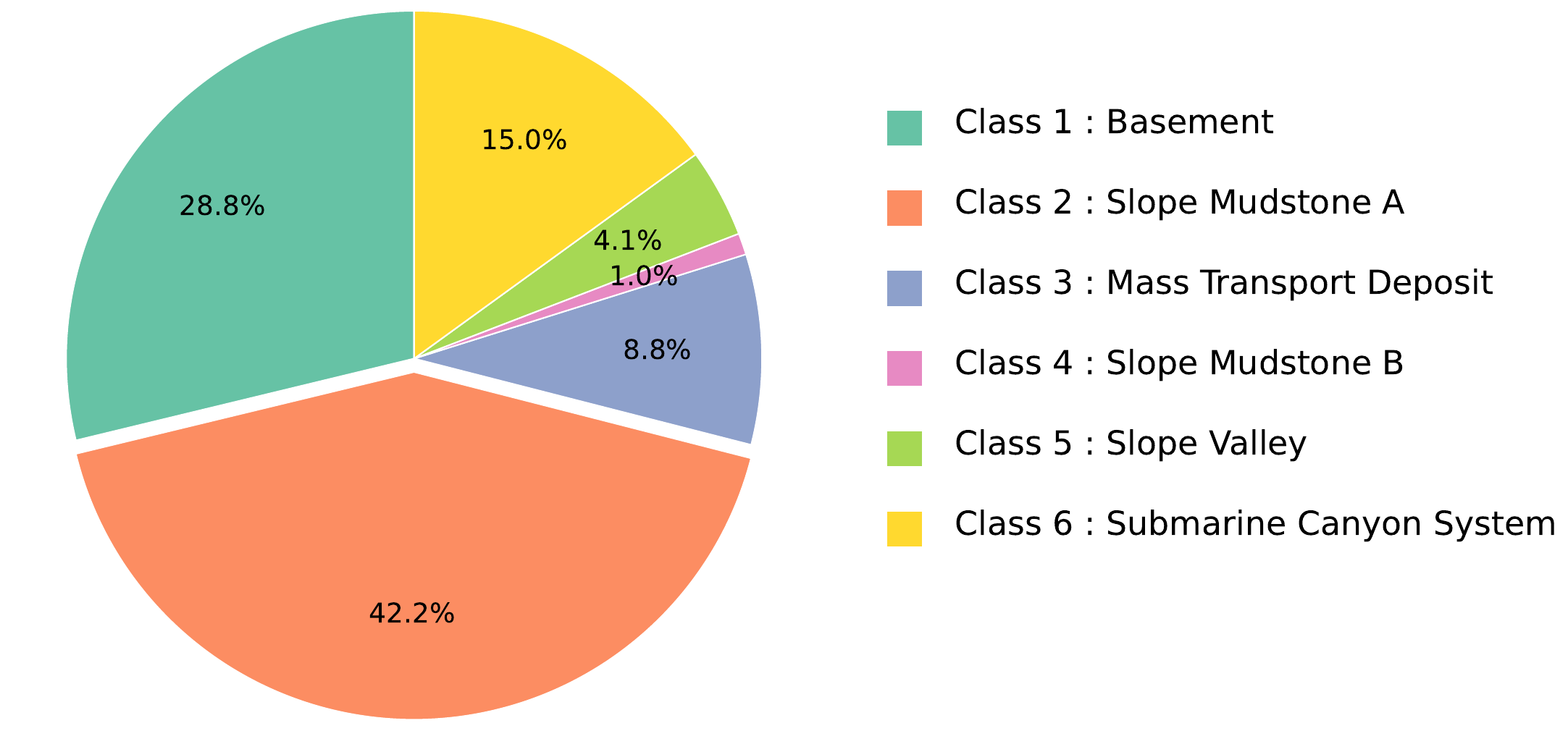}
    \caption{Parihaka dataset}
  \end{subfigure}
\caption{Distribution of facies classes from the two datasets.}
\label{fig:data}
\end{figure}

The F3 Netherlands dataset \cite{alaudah2019benchmark} is one of the most extensively studied public 3D seismic volumes and has become a standard benchmark in both seismic interpretation and facies classification. Acquired in the Dutch sector of the North Sea, the dataset provides high-quality and time-migrated seismic data covering approximately $384~\text{km}^2$. 
Its relatively clean signal, well-defined geological setting, and publicly accessible annotations have contributed to its widespread adoption in geoscience and data-driven interpretation studies. While numerous interpretations of the F3 dataset have been proposed in the facies classification task (\cite{alaudah2019benchmark}, \cite{ conocophillips2017malenov}, \cite{silva2019netherlands}), we adopt the same interpretation proposed by Alaudah et al. \cite{alaudah2019benchmark} (with a data shape of 601$\times$901$\times$255 voxels) to maintain consistency with the standard training pipeline of 2D seismic facies segmentation benchmark \cite{principled_benchmark_seismic_segmentation}. 
Figure \ref{fig:data}a then illustrates the distribution of its six facies classes.

\paragraph{Parihaka dataset}
The Parihaka dataset \cite{parihaka} represents another high-quality public 3D seismic dataset to perform facies interpretation and segmentation. The original survey was conducted offshore Taranaki, New Zealand
and the labels were then provided by Chevron USA \cite{parihaka}.
This dataset is highly valued thanks to its geological complexity, notably featuring a variety of faults and distinct gas hydrate deposits. Parihaka also includes 
six distinct geological facies classes as shown in Figure \ref{fig:data}b.
The complete raw data was defined by 841 inlines, 1116 crosslines, and 1006 time dimensions, but the official corresponding labels for some test regions remain unavailable, we can only use the training set (inlines 1 to 590 and crosslines 1 to 781) to perform our benchmark, as suggested in the partitioning protocol in \cite{principled_benchmark_seismic_segmentation}.


\paragraph{Data preparation process}

Following the standardized evaluation protocol established in \cite{principled_benchmark_seismic_segmentation}, our 3D facies segmentation benchmark adopts the five-fold cross-validation strategy with an 80/20 partitioning scheme. This setup is chosen not only because it is widely adopted in the seismic facies interpretation literature \cite{principled_benchmark_seismic_segmentation}, but also because it ensures fair and unbiased performance evaluation, while enabling statistical significance testing across independent data folds.

As this study focuses on 3D voxel segmentation, we follow prior benchmarks in medical image segmentation \cite{segmamba} and uniformly sample the seismic volumes into fixed-size sub-volumes of 128$\times$128$\times$128 voxels. Moreover, due to the limited spatial extent and irregular shape of the seismic annotated datasets, we apply mirror padding and use a stride of 64 when splitting each data fold into 3D sub-volume cubes. As a result, the F3 dataset, we obtain a total number of 975 and 195 sub-volumes for training and validation, respectively. For the Parihaka dataset, the corresponding numbers are 1215 and 405, as summarized in Table \ref{tab:data_pre}.



\begin{table*}[t]
\centering
\caption{Summary of the two studied datasets.}
\begin{tabular}{llccccc}
\toprule
Dataset & Location & Characteristics & Data shape & Fold shape & Train/Val samples \\
\midrule
F3 & North Sea, Netherlands & Shallow marine, moderate structural variation & 601$\times$901$\times$255 & 601$\times$180$\times$255 & 975 / 195 \\
Parihaka & Taranaki Basin, New Zealand & High structural complexity: faults, gas hydrates & 590$\times$781$\times$1006 & 590$\times$156$\times$1006 & 1215 / 405 \\
\bottomrule
\end{tabular}
\label{tab:data_pre}
\end{table*}

\paragraph{Evaluation Metrics}
To quantify 3D facies segmentation accuracy, we adopt the Dice coefficient, a widely-used metric for volumetric segmentation tasks.
For each semantic class, let $G_i$ and $P_i$ denote the ground-truth and predicted voxel values. 
The Dice coefficient is formulated as:

\begin{equation}
\text{Dice}(G, P) = \frac{2 \sum_{i=1}^{I} G_i P_i}{\sum_{i=1}^{I} G_i + \sum_{i=1}^{I} P_i},
\end{equation}
where $I$ denotes the total number of voxels in the 3D volume. The Dice coefficient ranges from 0 to 1, with higher values indicating better agreement between the predicted and ground-truth facies labels.



\subsection{Model architectures}


In this section, we present the implementation details of the three representative families of modern voxel-based 3D segmentation architectures, each reflecting a distinct design philosophy: a CNN-based method (3D-Unet \cite{3D_UNET}), a transformer-based model (UNETR \cite{unetr}), and a state-space-model-based (SegMamba \cite{segmamba}), evaluated for seismic facies segmentation.

\paragraph{3D-Unet}
    Inspired by the standard U-Net, the 3D-Unet model \cite{3D_UNET} consists of an analysis path (encoder) and a synthesis path (decoder). The encoder contains five stages, each composed of two $3\times3\times3$ convolutional layers followed by batch normalization and ReLU activation. Downsampling is performed using $2\times2\times2$ max-pooling layers applied after the first four encoder stages. The number of feature channels increases progressively from $16$, $32$, $64$, and $128$, with the bottleneck stage also operating at $128$ channels. The decoder mirrors the encoder by progressively increasing spatial resolution using trilinear upsampling. At each decoding stage, the upsampled feature maps are concatenated with the corresponding encoder features via the skip connections. Each decoder block again uses two $3\times3\times3$ convolutional layers with batch normalization and ReLU activation. The number of channels decreases symmetrically through $64$, $32$, and $16$ feature channels across the decoder. A final $1\times1\times1$ convolution maps the resulting $16$-channel feature volume to the desired number of output classes, which is $6$ facies in both F3 and Parihaka dataset.

    Specifically, unlike the original 3D U-Net \cite{3D_UNET} architecture, we do not apply downsampling below the final layer, while all upsampling steps fully restore the spatial dimensions. As a result, the output preserves the same spatial resolution as the input data, producing a complete $128\times128\times128$ voxel segmentation prediction. This ensures that each predicted voxel is aligned directly with the input while still benefiting from the hierarchical multi-scale features extracted throughout the encoder–decoder structure.

\paragraph{UNETR}

    Following UNETR \cite{unetr}, we implement UNETR, a 3D medical image segmentation model that combines a transformer-based encoder with a convolution-based decoder. Firstly, the 3D input volume is divided into uniform non-overlapping smaller 3D patches, which are then linearly projected into a K-dimensional embedding space and concatenate with learnable positional embeddings. Then, the transformer encoder processes this sequence via multi-head self-attention and MLP layers to capture long-range dependencies while preserving spatial information. Similiar to standard U-Net architecture, the features from multiple transformer layers are reshaped back into 3D tensors and directly merged with the decoder through skip connections. At each stage, 3×3×3 convolutions and deconvolutional layers are applied to upsample and fuse features, finally producing voxel-wise semantic predictions using a 1×1×1 convolution with softmax activation. This design enables UNETR to leverage both global context from transformers and local detail from convolutional operations, achieving effective performance in 3D segmentation.

\paragraph{SegMamba}
To the best of our knowledge, SegMamba \cite{segmamba} is currently the-state-of-art of 3D medical segmentation. This model composes of three main components: Gated Spatial Convolution (GSC), a 3D feature encoder with multiple Mamba blocks and a convolution-based 3D decoder for predicting segmentation maps with feature-level uncertainty estimation (FUE) to enhance feature reuse. 

Following  \cite{segmamba}, we adopt GSC to filter and preserve valuable characteristics of the input data using multiple 3D convolution layers. This component helps maintain the spatial relationships between voxels before flattening the 3D features into 1D sequences fed into the segmentation model. Specifically, GSC filters the input features by multiplied element-wise with 3D convolution kernels of sizes $3\times3\times3$ and $1\times1\times1$. Additionally, subsequent convolution block further fuses the features, and a residual connection is applied to the original input. For a 3D input data $x$, the GSC is given as:
\[
\mathrm{GSC}(x) = x + C_{3\times3\times3} \left( C_{3\times3\times3}(x) \cdot C_{1\times1\times1}(x) \right),
\]
where $x$ denotes the original 3D input volume. These features are fed into TSMamba blocks of the encoder. Moreover, by integrating downsampling within the TSMamba blocks, SegMamba maintains multi-scale representations while avoiding excessive sequence length, a limitation in standard transformer-based approaches like UNETR \cite{unetr}. Then, the decoder progressively fuses features from different scales using 3×3×3 convolutions and deconvolutions, guided by FUE modules that weigh features based on their uncertainty, producing accurate voxel-wise segmentation maps.




\section{Experiments and Results}
\label{sec:experiments}

In this section, we present the experimental setup and the comparative evaluation of the three 3D segmentation methods on each of the two studied datasets, together with an analysis of their model complexity. We further conduct a sensitivity analysis with respect to the size of the sub-volumes used as model inputs. All experiments are conducted using identical training protocols to ensure a fair and unbiased comparison.

\subsection{Experimental Setup}
We perform and compare each proposed method on two studied benchmarks. 
After completing each fold, we evaluate the models on the its corresponding validation set and report the mean accuracy w.r.t. each class and the average of accuracy of the whole dataset. 

\textit{Training Configuration:
}All experiments were performed on an NVIDIA A100 with 80 GB  of VRAM. The experiments were carried out on Ubuntu Server 18.04.3 LTS. To ensure comparability and reproducibility, the choice of hyperparameters was primarily followed by prior studies. Specifically, we adopted the hyperparameter settings commonly used in standard 3D segmentation frameworks \cite{3D_UNET, unetr, segmamba}, including learning rate schedules, optimizer configurations, and batch sizes. This approach not only facilitates direct comparison with existing methods but also ensures stable convergence and reliable performance across different experimental setups.

\subsection{Facies Segmentation Performance}



Table \ref{tab:results} reports per-class Dice scores and average performance for the three evaluated 3D segmentation methods on the Netherlands F3 and Parihaka datasets. Let us first compare the overall performance across the two datasets. On the F3 dataset, all methods achieve higher accuracy, reflecting the relatively clean signal and well-defined geological structures. In contrast, performance on the Parihaka dataset is consistently lower across all models (by approximately $10\%$), reflecting its geological complexity as described in Section \ref{sec:dataset_metric}. We note that a similar behavior has been also observed in prior 2D facies segmentation benchmarks \cite{principled_benchmark_seismic_segmentation}.

\begin{figure*}[t]
\centering
\begin{tabular}{cccc}
\includegraphics[width=3.75cm]{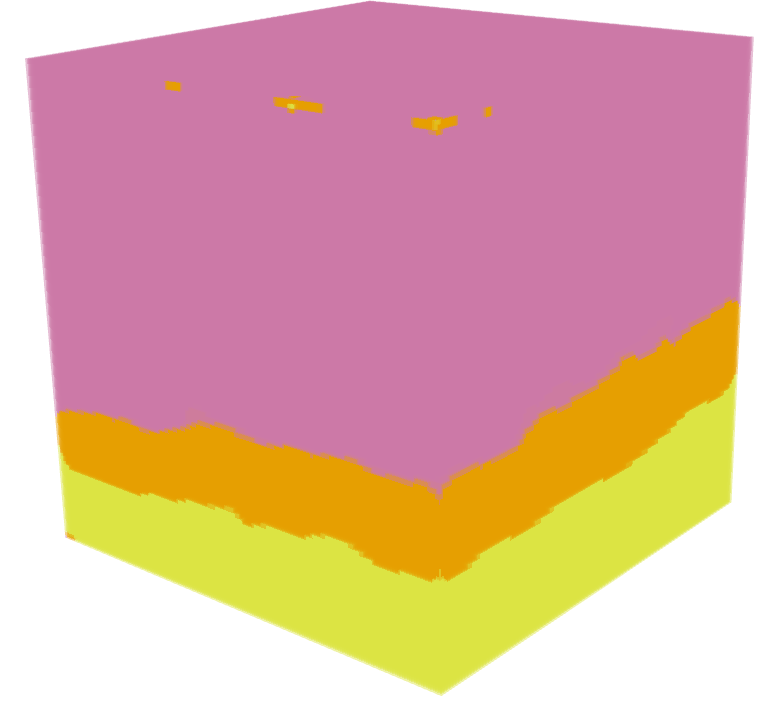} &
\includegraphics[width=3.75cm]{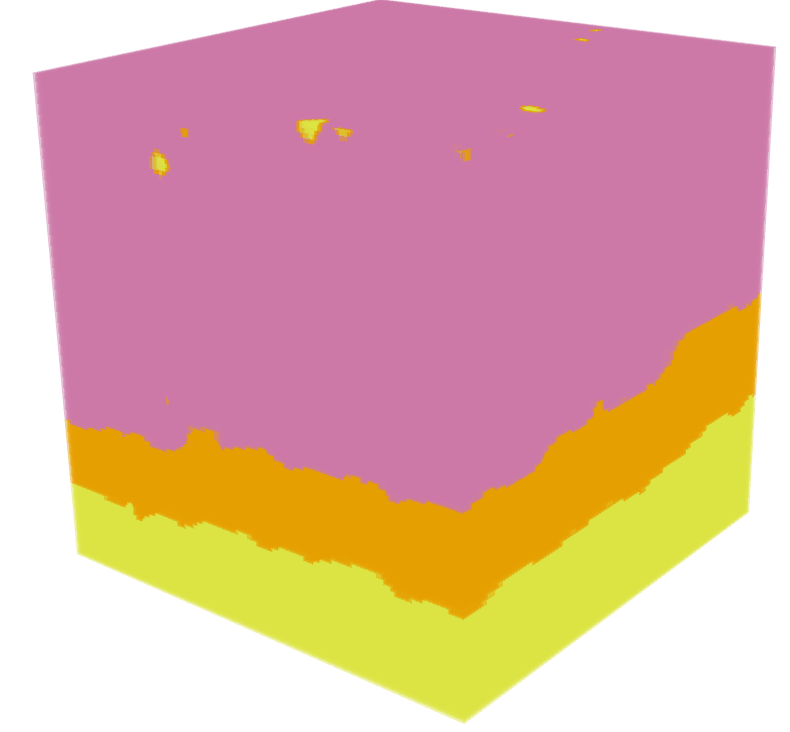} &
\includegraphics[width=3.75cm]{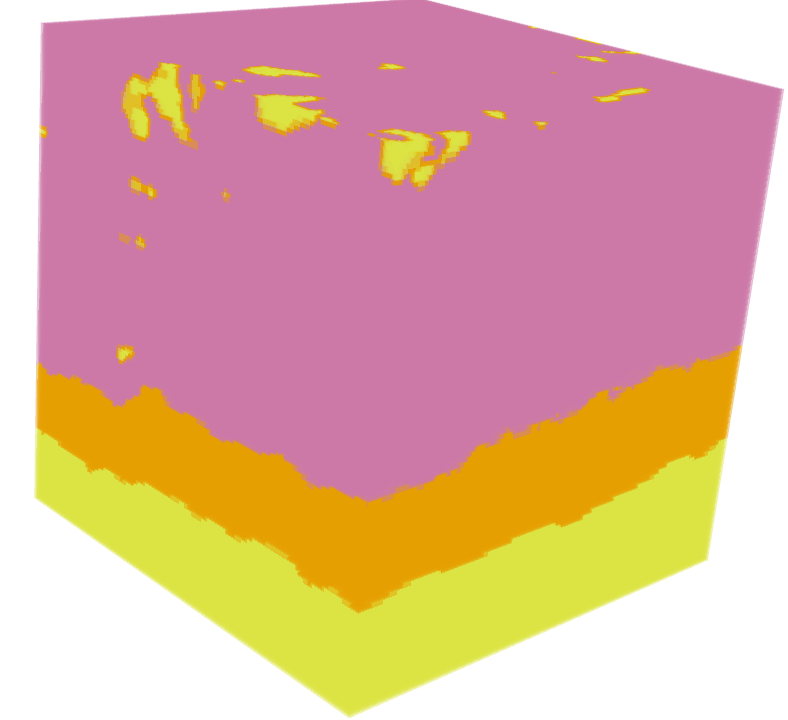} &
\includegraphics[width=3.75cm]{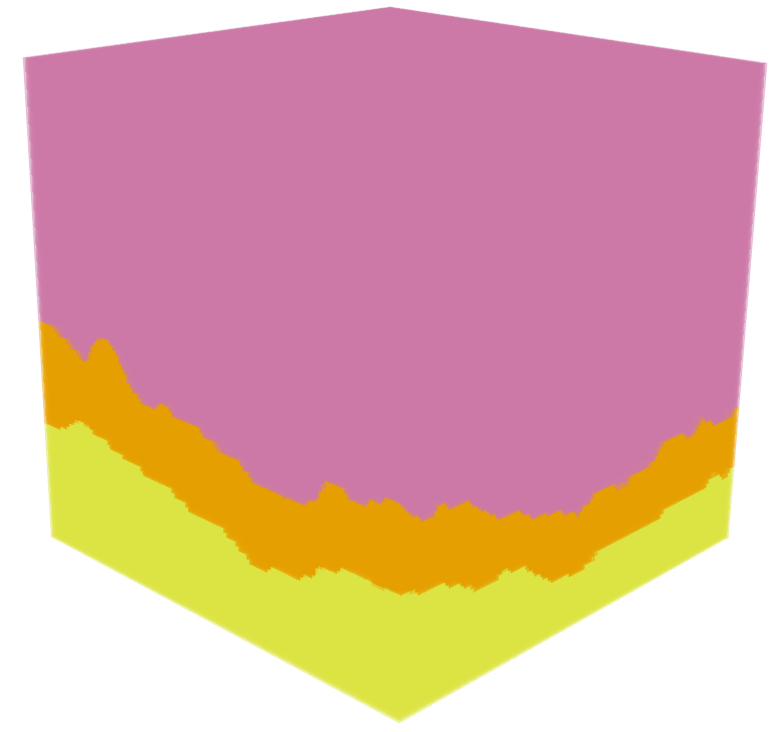}\\
\includegraphics[width=3.75cm]{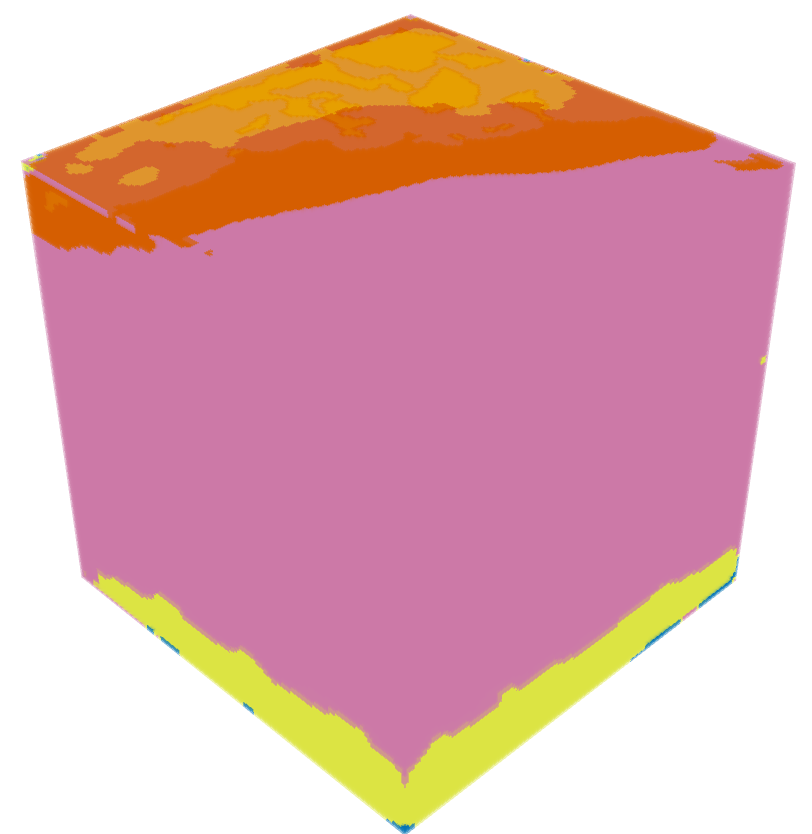} &
\includegraphics[width=3.75cm]{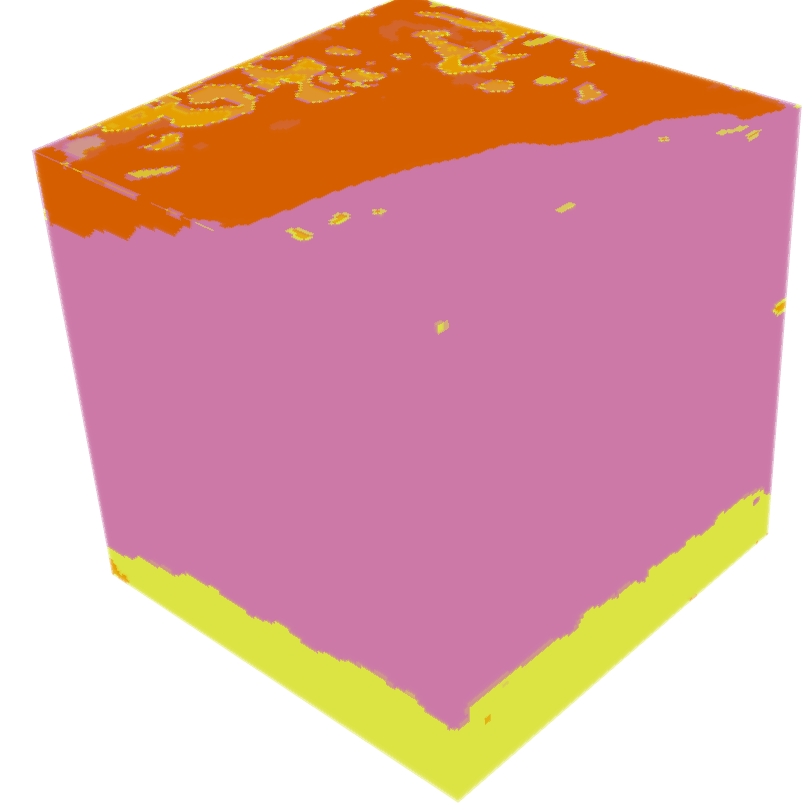} &
\includegraphics[width=3.75cm]{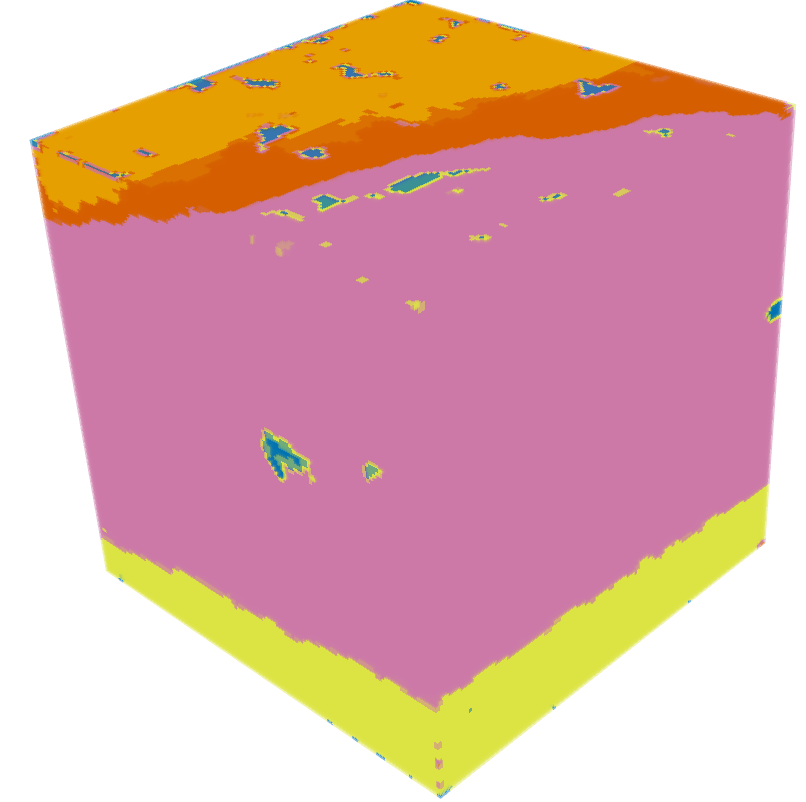}&
\includegraphics[width=3.75cm]{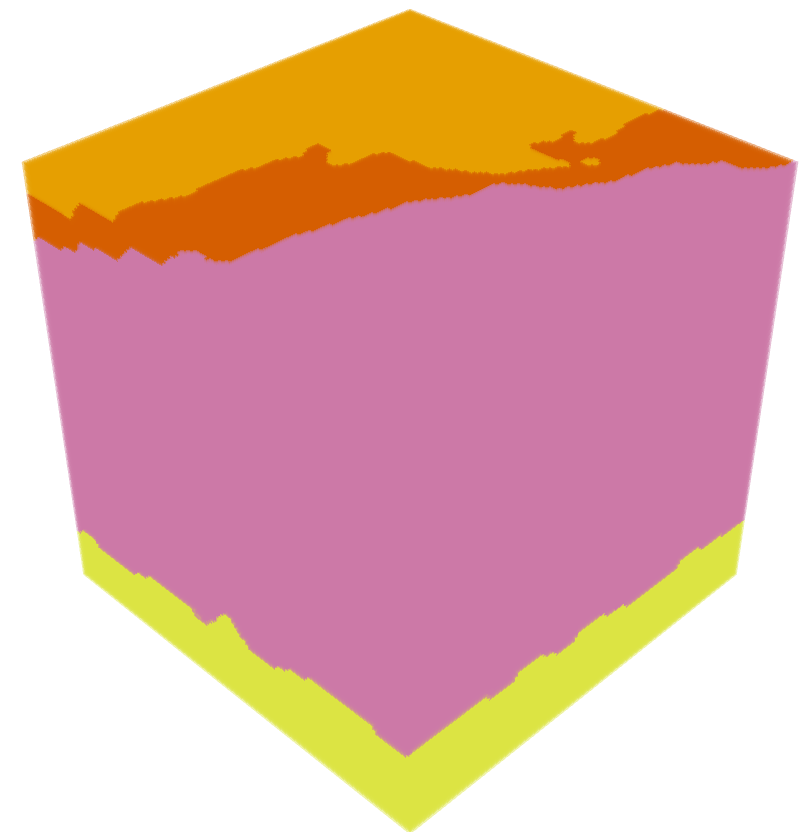} \\
3D-Unet & UNERT & SegMamba & Ground truth
\end{tabular}
\caption{Qualitative comparison of facies segmentation results obtained by the three methods.}
\label{fig:seg results}
\end{figure*}

Among the evaluated models, 3D U-Net achieves the highest average Dice score ($87.69\%$ for F3 and $75.01\%$ for Parihaka), demonstrating the effectiveness of convolutional architectures in capturing local volumetric features, particularly in scenarios with limited labeled seismic data.
Among the other two, SegMamba performs competitively but remains approximately $3\%$ lower than 3D UNet on both datasets. In contrast, UNERT shows comparatively weaker performance, suggesting that transformer-based architectures may struggle under data-scarce conditions and high computational demands in 3D seismic facies segmentation. In Figure \ref{fig:seg results}, we illustrate a qualitative comparison of the 3D facies segmentation results obtained by the three methods on two sample volumes from the F3 dataset. As observed, all models are able to recover the main facies, but 3D U-Net yields less noise compared to the other two. 
To this end, while UNETR and SegMamba have demonstrated noticeable effectiveness in 3D medical image segmentation \cite{segmamba}, the superior performance of 3D U-Net on both studied datasets suggests that convolutional inductive biases remain highly effective when geological structures exhibit strong local continuity and limited long-range dependencies. 


\begin{table}[h!]
\begin{subtable}{\linewidth}
\centering
\resizebox{\linewidth}{!}{
\begin{tabular}{lccccccc}
\hline
Model & Class 1 & Class 2 & Class 3 & Class 4 & Class 5 & Class 6 & Avg \\
\hline
3D Unet    & \textbf{96.67} & \textbf{88.02} & \textbf{96.4}  & 88.09 & \textbf{77.07} & 79.06 & \textbf{87.69} \\
UNETR      & 93.08           & 73.08           & 95.4           & \textbf{88.84} & 75.8            & 72.72 & 83.16 \\
SegMamba   & 96.05           & 75.94           & 95.97          & 80.29 & 75.43           & \textbf{81.57} & 84.21 \\
\hline
\end{tabular}}
\caption{Results on the F3 dataset}
\label{tab:result F3}
\end{subtable}
\vspace{1mm}

\begin{subtable}{\linewidth}
\resizebox{\linewidth}{!}{
\begin{tabular}{lccccccc}
\hline
Model & Class 1 & Class 2 & Class 3 & Class 4 & Class 5 & Class 6 & Avg \\
\hline
3D Unet   & 77.96 & 52.42 & \textbf{68.85} & \textbf{89.20} & \textbf{77.01} & \textbf{84.62} & \textbf{75.01} \\
UNETR      & 66.32 & 50.10 & 59.04 & 85.40 & 76.40 & 74.37 & 68.60 \\
SegMamba   & \textbf{81.88} & \textbf{72.19} & 70.11 & 69.66 & 59.53 & 81.23 & 72.44 \\
\hline
\end{tabular} }
\caption{Results on the Parihaka dataset}
\label{tab:result Parihaka}
\end{subtable}

\caption{Performance of the three methods for voxel-based 3D facies segmentation on two studied datasets. The best results are shown in \textbf{bold}.}
\label{tab:results}
\end{table}

\subsection{Model Complexity and Computational Cost}
In addition to the segmentation performance comparison, we analyze the model complexity and computational cost in Table \ref{model_complex_tab}. It is remarkable to note that 3D U-Net again outperforms the other two methods in computational efficiency. As a result, it achieves competitive performance while requiring significantly fewer parameters ($2.51$M), the lowest training memory ($5.79$ GB), and only $95.41$ GFLOPs. These results highlight that 3D U-Net constitutes a strong and reliable baseline for volumetric 3D seismic facies segmentation tasks.

\begin{table}[h]
\centering
\resizebox{\columnwidth}{!}{
\begin{tabular}{lrrr}
\hline
Model & Parameters (M) & Train Memory (GB)  & GFLOPs  \\
\hline
3D Unet   & 2.51   & 5.79   & 95.41    \\
UNETR     & 146.60  & 24.92  & 1868.45  \\
SegMamba  & 67.36  & 33.44  & 1427.20   \\
\hline
\end{tabular}}
\caption{Comparison of model complexity and computational cost. 
}
\label{model_complex_tab}
\end{table}

\subsection{Sensitivity Analysis on Sub-volume Size}

To investigate the impact of input resolution on segmentation performance, we conduct a sensitivity analysis using 3D U-Net with different sub-volume sizes on the F3 dataset. The results are summarized in Table \ref{tab:input size}.
Among the tested configurations, a sub-volume size of $128\times128\times128$ voxels
yields the best average Dice score of $87.69\%$, outperforming both smaller ($96\times96\times96$)
and larger
($144\times144\times144$)
input sizes. Smaller sub-volumes fail to capture sufficient spatial context, leading to degraded performance, while larger sub-volumes do not provide additional performance gains and may introduce optimization difficulties due to increased memory usage and reduced batch sizes.
These findings justify the choice of 
($128\times128\times128$)
as a balanced input resolution that effectively captures volumetric context while maintaining computational efficiency.
 
\begin{table}[h]
\resizebox{\linewidth}{!}{%
\centering
\begin{tabular}{lccccccc}
\hline
Input size & Class 1 & Class 2 & Class 3 & Class 4 & Class 5 & Class 6 & Avg \\
\hline
$96^3$  & 84.39 & 85.60 & 86.65 & 85.54 & 76.86 & \textbf{82.76} & 83.63 \\
$128^3$ & \textbf{96.67} & \textbf{88.02} & \textbf{96.40} & \textbf{88.09} & \textbf{77.07} & 79.06 & \textbf{87.69} \\
$144^3$ & 92.89 & 84.54 & 96.14 & 82.67 & 74.21 & 79.77 & 85.04 \\
\hline
\end{tabular}
}
\caption{Accuracy (\%) of 3D U-Net with different training data shapes on the Netherlands F3 dataset using five-fold cross-validation. The best results are shown in \textbf{bold}.}
\label{tab:input size}
\end{table}

\section{Conclusion}
\label{sec:conclusion}

We have presented the first comparative benchmark for 3D seismic facies segmentation using state-of-the-art voxel-based deep learning architectures. While existing facies analysis has traditionally relied on manual interpretation or 2D slice-based learning models, such approaches inherently fail to capture the volumetric continuity and spatial coherence of geological structures present in seismic data. By evaluating representative CNN-based, transformer-based, and state-space-based models on the two widely-used Netherlands F3 and Parihaka seismic datasets under standardized experimental settings, we have provided quantitative insights into their segmentation performance, computational complexity, and sensitivity to input shape.
Our results show that CNN-based architectures remain strong and competitive baselines for seismic facies segmentation. In particular, 3D U-Net consistently achieved the best overall performance on both datasets, while also exhibiting the lowest computational and memory cost among the evaluated methods. This indicates that convolutional operations are well suited for seismic volumes with strong local continuity and limited training data. In contrast, more complex architectures such as UNETR and SegMamba are underperformed 3D U-Net, highlighting the challenges of applying such data-hungry models in data-scarce seismic settings.




\small
\bibliographystyle{IEEEtran}
\bibliography{bibliography}

\end{document}